\documentclass[conference]{IEEEtran}
\IEEEoverridecommandlockouts
\usepackage{silence}
\usepackage{silence}
\usepackage{cite}
\usepackage{booktabs,makecell}
\usepackage{subcaption}
\usepackage{amsmath,amssymb,amsfonts}
\DeclareMathOperator{\E}{\mathbb{E}}
\usepackage[dvipsnames]{xcolor}
\usepackage{graphicx}
\usepackage{textcomp}
\usepackage{multirow}
\usepackage{xcolor}
\usepackage{hyperref}
\usepackage{neuralnetwork}
\usepackage{array}
\usepackage{dblfloatfix}
\newcolumntype{P}[1]{>{\centering\arraybackslash}p{#1}}
\newcolumntype{M}[1]{>{\centering\arraybackslash}m{#1}}

\hypersetup{%
  colorlinks=true,%
  linkcolor={blue},
  citecolor={blue!65!blue},
  urlcolor={blue!80!blue}
  bookmarksnumbered=true,
  bookmarksopen=true}

\def\BibTeX{{\rm B\kern-.05em{\sc i\kern-.025em b}\kern-.08em
    T\kern-.1667em\lower.7ex\hbox{E}\kern-.125emX}}
\begin{document}

\title{RaceGAN: A Framework for Preserving Individuality while Converting Racial Information for Image-to-Image Translation\\

}

\author{\IEEEauthorblockN{Mst Tasnim Pervin\IEEEauthorrefmark{1},
George Bebis\IEEEauthorrefmark{2}, Fang Jiang\IEEEauthorrefmark{3} and
Alireza Tavakkoli\IEEEauthorrefmark{4}}
\IEEEauthorblockA{\IEEEauthorrefmark{1}\IEEEauthorrefmark{2}\IEEEauthorrefmark{4}Department of Computer Science \& Engineering,\IEEEauthorrefmark{3}Department of Psychology,\\
University of Nevada, Reno, USA\\
Email: \IEEEauthorrefmark{1}mpervin@unr.edu,
\IEEEauthorrefmark{2}bebis@unr.edu,
\IEEEauthorrefmark{3}fangj@unr.edu,
\IEEEauthorrefmark{4}tavakkol@unr.edu}}

\maketitle

\begin{abstract}
Generative adversarial networks (GANs) have demonstrated significant progress in unpaired image-to-image translation in recent years for several applications. CycleGAN was the first to lead the way, although it was restricted to a pair of domains. StarGAN overcame this constraint by tackling image-to-image translation across various domains, although it was not able to map in-depth low-level style changes for these domains. Style mapping via reference-guided image synthesis has been made possible by the innovations of StarGANv2 and StyleGAN. However, these models do not maintain individuality and need an extra reference image in addition to the input. Our study aims to translate racial traits by means of multi-domain image-to-image translation. We present RaceGAN, a novel framework capable of mapping style codes over several domains during racial attribute translation while maintaining individuality and high level semantics without relying on a reference image. RaceGAN outperforms other models in translating racial features (i.e., Asian, White, and Black) when tested on Chicago Face Dataset. We also give quantitative findings utilizing InceptionReNetv2-based classification to demonstrate the effectiveness of our racial translation. Moreover, we investigate how well the model partitions the latent space into distinct clusters of faces for each ethnic group.

\end{abstract}


\section{Introduction}

Generative adversarial networks are a special type of artificial intelligence algorithm employed in generative machine learning. Since their introduction by Ian Goodfellow et al. in 2014, GANs have drawn a lot of interest because of their capacity to produce realistic synthetic data for several tasks, including image generation, image manipulation and transformation, image-to-image translation, and even domain conversion, such as image-to-text generation or vice-versa \cite{ian2014generative}. GANs consist of two basic components, i.e., the generator and the discriminator. Both of these modules are neural networks, where generator tries to generate artificial data by taking inspiration from input data and discriminator seeks to distinguish between real inputs and generator's production. The discriminator acts as a binary classifier, identifying whether the generated data is real or fake. This leads to a competitive adversarial training process for both the counterparts, where the generator learns to generate realistic data identical enough to the real data to fool the discriminator, while the discriminator learns to enhance its ability to detect the fake generated data. This adversarial training plays a vital role in achieving stability.


By enabling users to provide conditional transformation, Conditional GANs (cGANs) have increased the possibilities of image-to-image translation \cite{mirza2014conditional}. This provides additional flexibility in the translation process by enabling a more regulated approach to image generation. The effectiveness of this model lies in its ability to use conditional generation and adversarial training to generate visually pleasing outcomes. Pix2Pix, an improvised version of conditional GAN, \cite{mirza2014conditional}\cite{isola2017image} employs paired datasets for training in a supervised learning approach where the generator tries to do image-to-image translation from the source domain to the target domain. One of the fundamental features of Pix2Pix is the use of paired data for training, where each input image in the source domain is matched with its counterpart image in the target domain. By using supervised learning to map the source and target domains directly, the quality of the generated images is increased. 

While the models previously discussed are capable of handling image-to-image translation, their functionality varies largely depending on the application. For instance, Pix2Pix\cite{isola2017image} needs paired data to be available; the training data must comprise inputs in pairs from the source and target domains. It may create a street view from satellite photos or generate a scenic winter landscape from summer images by transforming images. However, it is not feasible to have matched data for both source and target domains of the same input image in all circumstances, such as gender or race conversion. By using an unpaired dataset, CycleGAN\cite{zhu2017unpaired} solves this issue; nevertheless, each pair of domains requires a different model to be trained. To address this problem, StarGAN\cite{choi2018stargan} uses a single model to manage numerous domain conversions. However, if the style of the input domain differs greatly from that of the target domain, the model will encounter large domain gaps and will have trouble understanding the intended mapping. To solve this issue StarGAN-v2\cite{choi2020stargan} includes two more modules—the mapping network and the style encoder—to extract different style codes utilizing input images and latent space. Nevertheless, this modification becomes reference dependent and eliminates the unique individuality of the input image by modifying the source domain image with the reference style from the target domain. 

Our goal is to change the domains of the input images using domain-specific information while preserving the individuality of the subject. If domain conversion can be done while manipulating low-level semantics, it could be really helpful to mitigate bias in the datasets to be applied for generative models. To accommodate this, we propose an image-to-image translation model for racial feature conversion using the architecture of StarGAN\cite{choi2018stargan} as its backbone. To facilitate domain-specific style extraction from latent space, we develop a style extractor module, drawing inspiration from the mapping network of StarGAN-v2\cite{choi2020stargan}. It enables the model's integration of the unique style of target domain while maintaining the individuality from the input source domain. The main contributions to this work are as follows:
\begin{itemize}
    \item Our model comprises a style extractor network to extract domain-specific styles and simultaneously trains for various domains with only a single generator/discriminator pair. 
    \item The proposed approach maps racial traits, e.g., demographic skin color, eye size and orientation, ratio of upper to lower lips, etc., from one domain to many target domains.
    \item The model maintains high-level semantics (e.g., hairstyle, cosmetics) and posture expressions.
    \item Our model preserves the individuality of the input images. 
    \item Finally, since the translation of a racial trait must influence the identity, we propose an InceptionResNetv2-based classification approach to evaluate whether the person's individual features is retained rather than relying on visual representations of the translated domain images alone.
    
    \end{itemize}


\begin{figure*}[ht!]
    \centering
    \includegraphics[width=14cm,height=4cm]{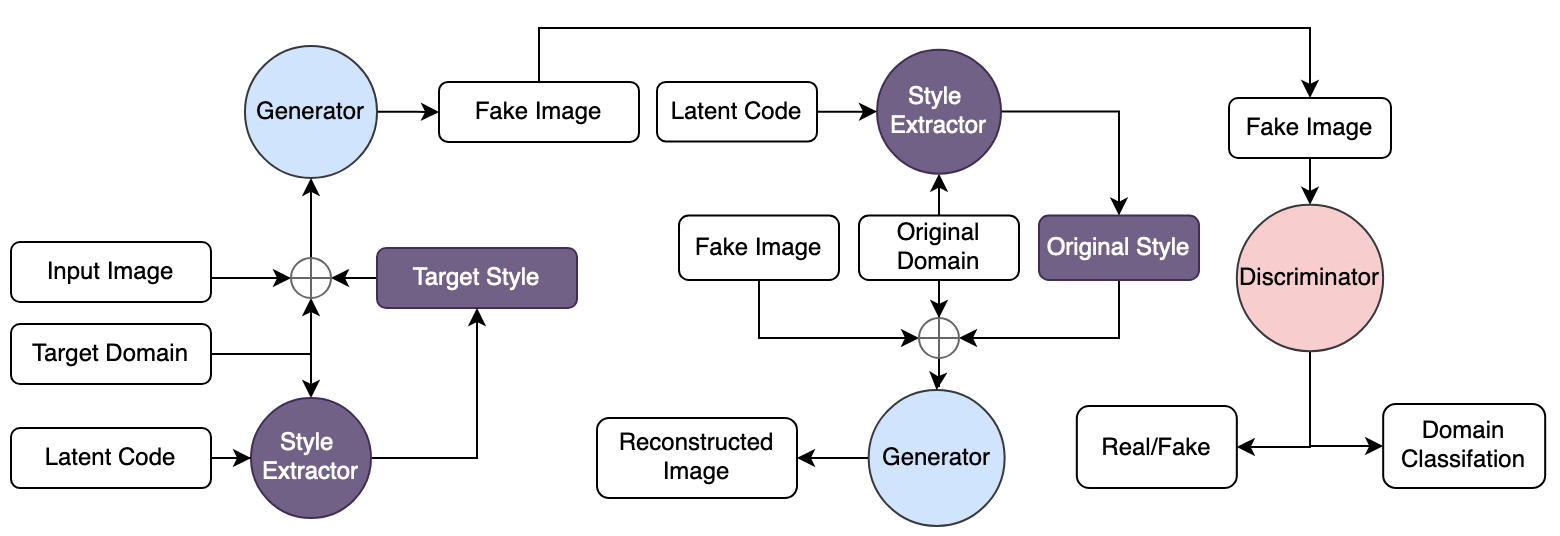}
    \caption{Overview of the proposed framework of RaceGAN, consisting of 3 modules: generator, style extractor, and discriminator. The style extractor generates target styles using latent code and the target domain label. Using real image and the depth-wise concatenation of target style and target domain label, the generator generates a fake image, which is used both for original input reconstruction and discrimination. The discriminator tries to discriminate between real and fake images and find the respective domain.}
    \label{flow}
\end{figure*}
\section{Related Works}
In the field of computer vision and image processing, Pix2Pix is a seminal work authored by Phillip Isola et al. \cite{isola2017image}, despite the limitation that it is dependent on paired datasets. Later on, a number of researchers investigated this concept with multiscale GAN architecture in a variety of applications, including semantic segmentation, edge guiding, image impainting, and more. In order to address the limitation, Jun-Yan Zhu et al. introduce the idea of cycle consistency \cite{zhu2017unpaired}. 
The term cycle consistency represents the concept that an image should return to its original form if it is translated from one domain to another and back again. Based on this idea of cycle consistency loss, the model can be more optimized for reconstructing images from the source domain. This contributes to the accuracy and realism of the translations. The study presents that the CycleGAN\cite{zhu2017unpaired} architecture is comprised of two jointly trained Pix2Pix GANs\cite{isola2017image}. The model is a cycle-consistent adversarial network made up of two discriminators and two generators. Cycle consistency loss provides extra oversight during training by guaranteeing that an image translated from one domain to another and back reconstructs the original image. Both of the models, CycleGAN\cite{zhu2017unpaired} and DiscoGAN\cite{kim2017learning}, preserve significant features between the translated image and the input. Nevertheless, these frameworks are limited to simultaneously learning the relationships between two distinct domains. Their methods are not very flexible when handling several domains because each pair of domains requires training a distinct model. In the paper UNIT\cite{liu2017unsupervised}, the use of unpaired data has also been investigated. They present a shared latent space, which is modeled after the CoGAN architecture \cite{liu2016coupled}. Images are encoded into a latent code for both domains without necessitating a direct mapping between the domain images. The paper StyleGAN\cite{karras2019style} examines more into the idea of latent space. StyleGAN uses the concept of neural style transfer to create incredibly realistic images by adjusting both high-level and low-level semantics. The model specifies a beginning point for image generation using latent space. In order to get an appearance or alter hair color, StyleGAN then blends many looks at different phases. In line with Progressive GAN\cite{karras2017progressive}, the model improves the picture quality by progressively increasing the resolution during training.
However, as its generator isn't made to take images as input, using this method to alter an actual image takes some effort. In contrast, StarGAN\cite{choi2018stargan}, uses a single generator and discriminator to do image-to-image translation tasks for various domains using real images as input. Through the utilization of mask vectors, the model was able to learn mappings across numerous datasets and domains. However, despite its remarkable effectiveness in multi-domain translation, StarGAN\cite{choi2018stargan} struggles to generate high-resolution images because it lacks progressive growth architecture and experiences mode collapse, which results in the production of only a limited number of distinct images.Even though they used the Wasserstein GAN with gradient penalty to enhance training objectives\cite{gulrajani2017improved}, the mode collapse problem remained unresolved. To overcome the restriction of diversity, the same authors later modified the model and created StarGAN-v2 \cite{choi2020stargan}. They included a mapping network and a style encoder as two additional modules to help the generator with reference-guided image synthesis. It also introduces a progressive growth based training process that allows better quality pictures to be produced. Even with these improvements, mode collapse and scalability problems may still arise, especially when dealing with large-scale datasets. Even so, it's a big step forward from its predecessor, providing image-to-image translation approach with more powerful and adaptable features.

\section{RaceGAN}
The primary goal of this research is to create a model capable of translating images between different racial domains. Inspired by StarGAN\cite{choi2018stargan}, we create a GAN architecture for multi-domain translation that consists of just one generator and one discriminator. We must extract race style from several race domains in order to translate racial traits. Inspired by the mapping network of StarGAN-v2\cite{choi2020stargan}, we propose a style extractor module to add style information relevant to racial domains from the latent space of several domains. This network extracts racial style codes, which help the generator convert the input image into a racially translated image of the target domain. In this section, we describe the overall architecture of the implemented network. Essential loss functions that have been incorporated for training are also described.

\subsection{Proposed Architecture}
The input images of a human face can be assumed to consist of content space and context space. The context space reflects the ethnic traits, whereas the content space represents the individual features and high-level semantics (such hair color, accessories, etc.). Our goal is to translate context space's racial traits while maintaining the input images' high-level semantics and individual identities. Let $X$ be the set of input images and $C$ be the racial domains of the input images (e.g., Asian, Black, or White) for easier comprehension. The racial domain $C$ and race-specific style codes $S$ would form the context space of $X$, while the personal style, including high-level semantics, would be the content space. Using domain-specific styles, $s \in S$, retrieved by the style extractor network, we train the generator to create pictures $y$ that correspond to target domains given an image $x \in X$ and matching label $c \in C$. All of the modules in the proposed architecture are shown in Figure \ref{flow}.

\textbf{Generator:}
The generator network, $G$, is trained to produce images that are exclusive to a certain racial domain by adjusting context space while maintaining the high-level characteristics of the input images, such as hair color, cosmetics, accessories, and style. To force certain kind of image generation in the target domain via $G(x,c,s) \xrightarrow{} y$, the network accepts as input images, $x$, target domain label, $c$, and style information of target domain, $s$ collected from style extractor network.

\textbf{Style Extractor:}
Our style extractor network, $E$, extracts style patterns, $s=E_c(z)$, where $c$ indicates the target domain, $z \in Z$ is latent code randomly sampled from a Gaussian normal distribution matching the real image size, and $E_c()$ represents the resultant style code for the corresponding target domain, $c$. The network is built of several multi-layer perceptrons consisting of multiple output branches dedicated to all possible domains. The network employs the label information to extract the desired style code for that specific domain only.
\begin{figure*}[ht!]
    \includegraphics[width=18cm,height=13cm]{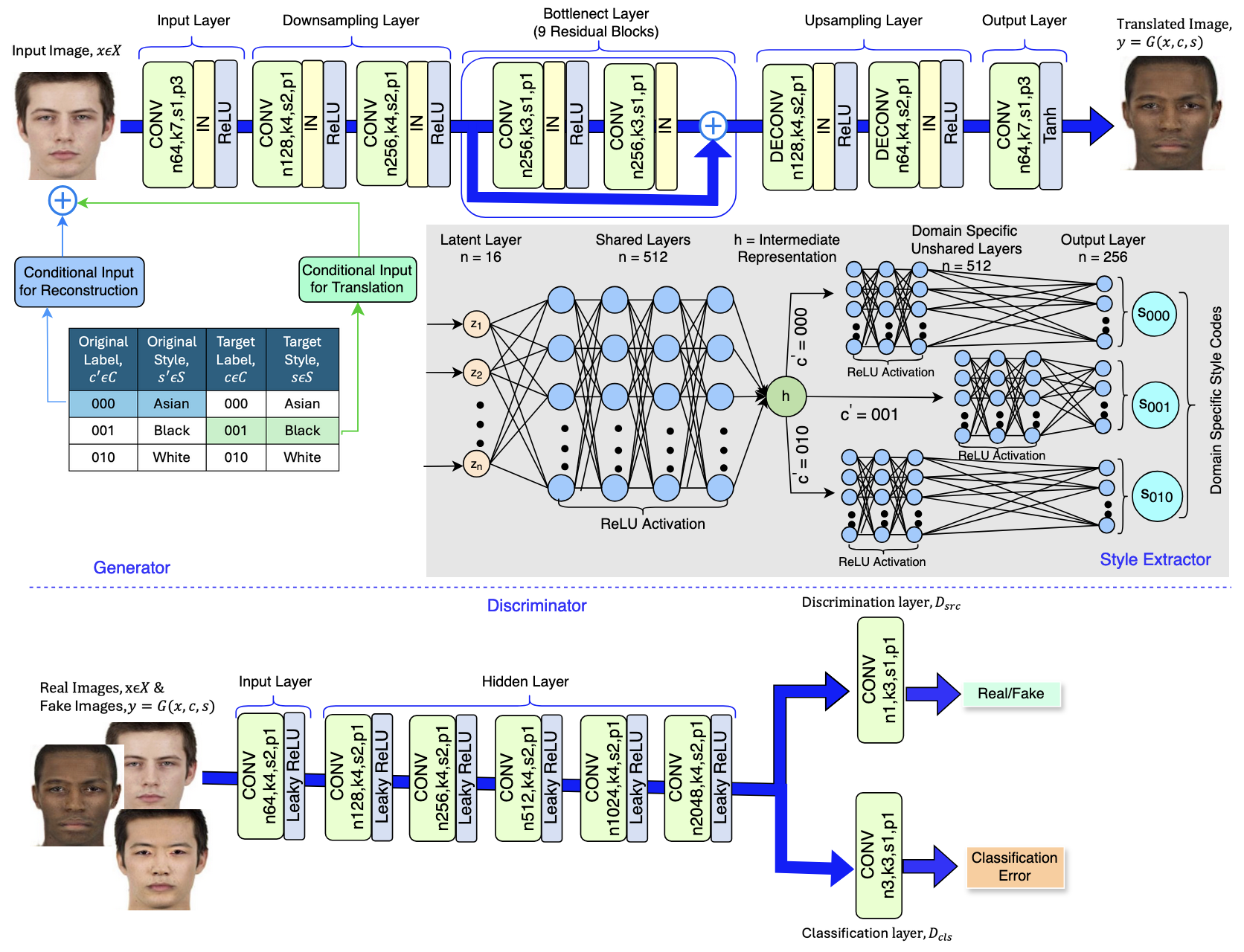}
    \caption{Network Architectures}\vspace{-0.2cm}
    \label{GMD}
\end{figure*}
\textbf{Discriminator:}
The discriminator network, $D$, consists of an auxiliary classifier that not only can discriminate among real and fake images generated from the generator but also can classify real and fake images to their corresponding domains, $D(x)\xrightarrow{}[D_{src}(x),D_{cls}(x)]$. The auxiliary classifier helps the discriminator control domains instead of using multiple discriminators for all domains individually. 
\subsection{Loss Functions}
For the input image, $x \in X$, respective domain label, $c^\prime \in C$, target domain label, $c \in C$, and corresponding target style, $s\in S$, the following objective functions help us to conduct the training.

\textbf{Adversarial loss:} The primary objective of the adversarial loss is the exact resemblance between the produced and original images. In order to create an image, first a random latent code $z$ and a random target domain $c$ are generated. These are then passed into the style extractor to generate style code $s$ for a particular domain. $G(x,c,s)$ generates images of various domains conditioned by domain labels and the style codes. Equation \ref{adv} represents the adversarial loss, which is built based on the WGAN-GP objective function\cite{gulrajani2017improved}. By employing gradient penalties, it contributes to more stable training. The primary advantage of the gradient penalty is that it penalizes the model and resolves the vanishing gradient problem if the gradient norm differs from the target norm value of 1\cite{gulrajani2017improved}. We must create an interpolated distribution $\widehat{x}$, evenly sampled from a generated image $y$ and a real image $x$, in order to put this into action. The WGAN article states that $\lambda_{gp}$, the penalty factor, is set to 10\cite{gulrajani2017improved}. Through the process of maximizing the objective function, the discriminator $D$ is able to identify the class or domain of the input by providing $D_{cls} (x)$, and it also attempts to distinguish between real and fake images by providing a probability, $D_{src} (x)$. The generator $G$ aims to minimize this objective in order to generate fake images that mimic real ones and trick the discriminator.
\begin{equation}
\begin{split}
\mathcal L_{adv} & = \E_{x}[D_{src}(x)] - \E_{x,c,s}[D_{src}(G(x,c,s)] \\
&  - \lambda_{gp} \E_{x}[(\|\nabla_{\widehat{x}}D_{src}(\widehat{x})\|_{2}-1)^{2}]
\end{split}
\label{adv}
\end{equation}
\begin{figure*}[b]
    \centering
    \includegraphics[width=18cm,height=7cm]{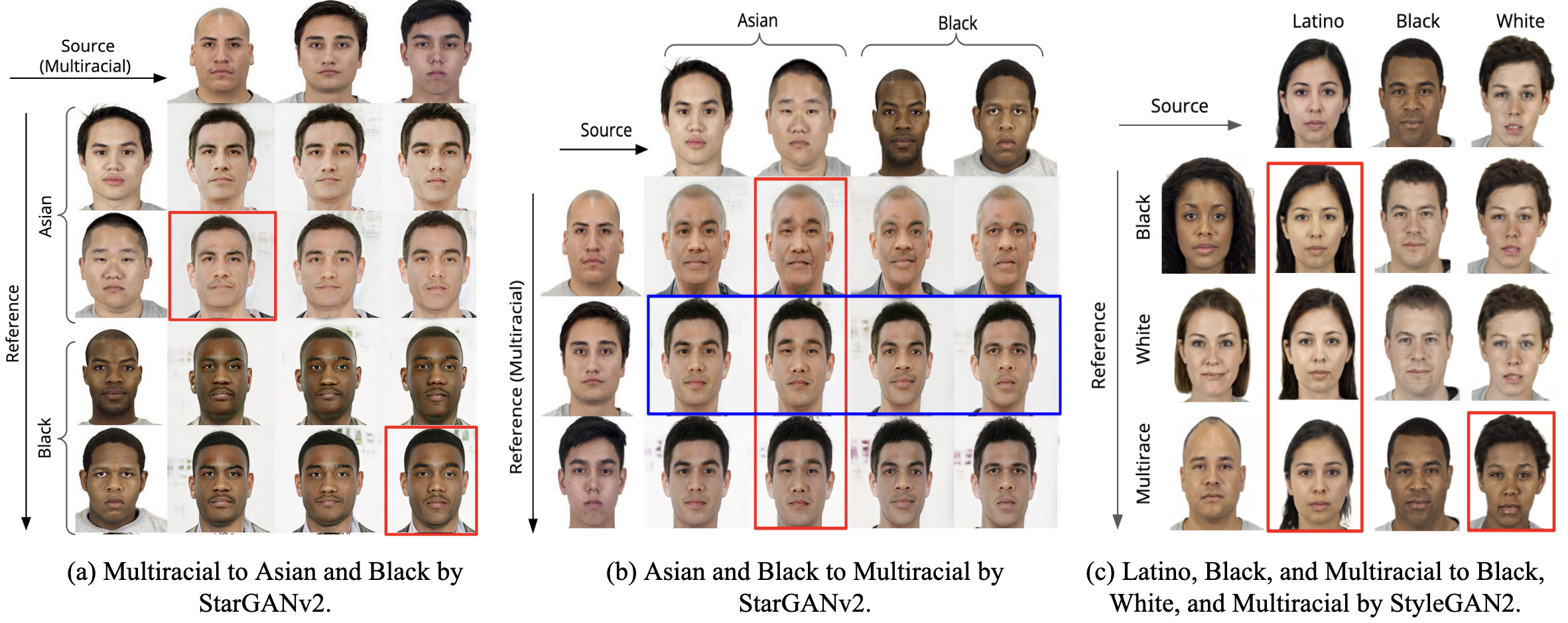}
    \caption{Issues StarGANv2 and StyleGAN2 face for reference-guided image translation using the CFD dataset: (a) High-level attributes (hair style, skin color) are transferred without preserving source individuality for the highlighted face in red. (b) The highlighted column in red share a single individuality from source with most low-level attributes (expression, pose, eye orientation) preserved and carrying style from reference but the row-wise synthesis (highlighted in blue) results in changing the person's individuality, (c) Some high-level attributes are transferred randomly for highlighted faces but the race conversions are incorrect.}
    \label{img_compare}
\end{figure*}
\textbf{Domain classification loss:} Since our goal is to generate images across several domains, it is imperative that we distinguish between authentic and fake images. As the last layer of the discriminator, $D$, a classifier module is linked to account for it. Our goal is to generate fake images, $y$, using the input image $x$, target domain label, $c$, and target domain style, $s$. To be clear about the particular features of each domain, we classify $y$ into its right domain with the use of an auxiliary classifier. Using domain classification loss, the discriminator, $D$, and the generator, $G$, are both tuned to perform faultless image classification of real and fake. The following Equation \ref{real_cls} is used to optimize $D$ using domain classification loss. During training, we are provided both the original domain label, $c^{\prime}$, and the input images, $x$. Using the estimated probability distribution $D_{cls}(c^{\prime}|x)$ and the help of real image classification into its proper domain $c^{\prime}$, D optimizes itself by minimizing $\mathcal L_{cls}^{real}$.
\begin{equation}
\begin{split}
\mathcal L_{cls}^{real} & = \E_{x,c^{\prime}}[-log D_{cls}(c^{\prime}|x)]
\end{split}
\label{real_cls}
\end{equation}
For optimizing G, we need to make it more efficient so that it can generate more domain specific realistic images. G tries to minimize the domain classification loss of its own generated fake images. This objective can be defined as Equation \ref{fake_cls}.
\begin{equation}
\begin{split}
\mathcal L_{cls}^{fake} & = \E_{x,c,s}[-log D_{cls}(c|G(x,c,s)]
\end{split}
\label{fake_cls}
\end{equation}

\textbf{Style reconstruction loss:} The generator is used twice to ensure correct image-to-image translation: once to create a fake image $y$ of a target domain $c$ from an actual image $x$ and a target domain style $s$. Subsequently, rebuild the input image using the fake image $y$, $c^{\prime}$, the input image's original domain, and $s^\prime$, the original domain style code. In line with earlier methods, we employ cycle consistency loss\cite{zhu2017unpaired} as in Equation \ref{cycle} to achieve this.
\begin{equation}
\begin{split}
\mathcal L_{rec} & = \E_{x,c,c^{\prime},s,s^{\prime}}[\| x - G(G(x,c,s),c^{\prime},s^{\prime})\|_{1}]
\end{split}
\label{cycle}
\end{equation}
To calculate reconstruction loss, we use $L1$ Distance. Previous objectives helped G generate realistic and domain-specific images, but now, with the goal of minimizing reconstruction loss, G can also preserve the maximum high-level features of real images to maintain individuality and control diversity. Eventually, by considering all of these loss functions, the generator $G$ and the discriminator $D$ are optimized by following Equations \ref{obj_G} and \ref{obj_D}, where $\lambda_{cls}$ and $\lambda_{rec}$ are weighting factors to control the effect of domain classification loss and style reconstruction loss, respectively. Following the implementation of StarGAN\cite{choi2018stargan}, $\lambda_{cls}$ has been set to 1 and $\lambda_{rec}$ has been set to 10.
\begin{equation}
\begin{split}
\mathcal L_{G} & = \mathcal L_{adv} + \lambda_{cls} \mathcal L_{cls}^{fake}+\lambda_{rec}\mathcal L_{rec}
\end{split}
\label{obj_G}
\end{equation}
\begin{equation}
\begin{split}
\mathcal L_{D} & = - \mathcal L_{adv} + \lambda_{cls} \mathcal L_{cls}^{real}
\end{split}
\label{obj_D}
\end{equation}

\section{Implementation Details}
\subsection{Dataset Description:}
In this study, the University of Chicago's Chicago Face Database (CFD) was employed \cite{ma2015chicago}. The dataset is solely meant to be used for research purposes and is accessible online upon request. It provides numerous usable extensions in addition to the primary image set. The dataset contains high-resolution images of men and women between the ages of 17 and 65 from various ethnic backgrounds. The primary collection, CFD, has 597 distinct images of self-identified male and female people from the United States who are of various racial backgrounds, including Asian, Black, White, and Latino. In addition to a few joyful, depressed, and scared emotions, this set mostly consists of neutral facial expressions. The CFD-MR\cite{ma2021chicago} and CFD-INDIA\cite{lakshmi2021india} extension sets have been developed for multiracial faces and Indian faces, respectively. There are 88 unique people with distinct facial expressions in CFD-MR, while for CFD-INDIA, the number is 142. We limited our first experimentation to images of White, Black, and Asian males with neutral facial expressions. We will also be including more variants in subsequent studies. These three classes are represented by the distribution of male and female faces: Asian (male:52, female:57), Black (male:231, female:295), and White (male:288, female: 236). This distribution makes it easy to identify the issue of class imbalance, which has a negative impact on the outcome overall.

\subsection{Network Architecture:}
The model architecture is inspired from the implementation of both StarGAN\cite{choi2018stargan} and StarGAN-v2\cite{choi2020stargan} adopting baseline from DIAT\cite{li2016deep}, CycleGAN\cite{zhu2017unpaired}, and ICGAN \cite{perarnau2016invertible}. We also try to do experimentation with StyleGAN2 \cite{karras2019style}. 

\textbf{Generator:}
Three layers make up the generator network: downsampling, bottleneck, and upsampling. The first convolutional layer with a stride of $1$ and a kernel size of $7\times7$ processes the three-channel input images. In the generator portion of Figure \ref{GMD}, the notation $IN$ denotes instance normalization. Downsampling is accomplished by two more convolutional layers with a kernel of $4\times4$ and a stride of $2$. A bottleneck made up of nine residual blocks with a kernel of $3\times3$ and a stride of $1$ follows the downsampling layer. Two transposed convolutional layers of the $4\times4$ kernel and stride $2$ make up the upsampling module. Instance normalization and the ReLU activation function were employed in each of the preceding hidden layers. After the upsampling layer, the feature maps are converted from the depth of $64$ to the depth of $3$ for the purpose of creating RGB fake images using an output convolutional layer consisting of kernel $7\times7$, stride $1$, and hyperbolic tangent activation function.
\begin{figure*}[!ht]
\centering
\begin{subfigure}{.33\textwidth}
  \centering
  \includegraphics[width=5.5cm, height=5cm]{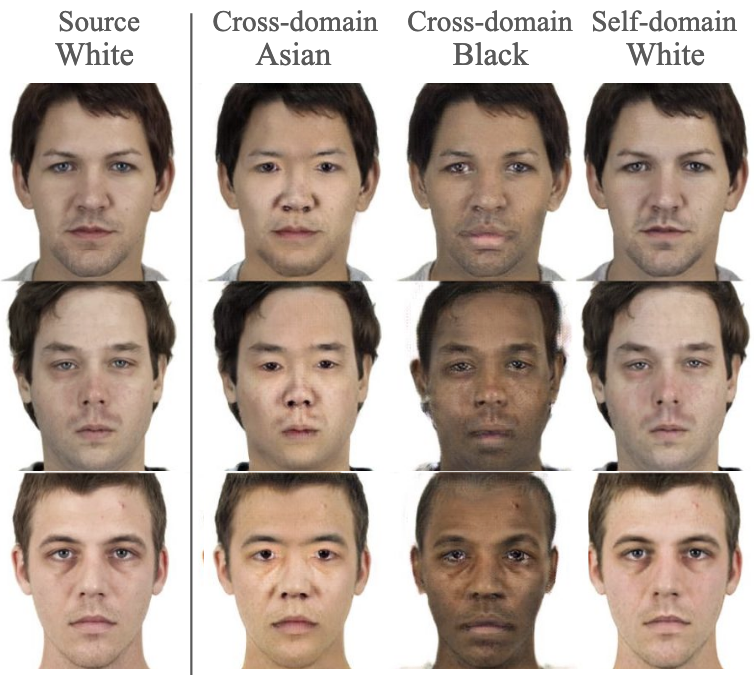}
  \caption{White to target domains}
  \label{racegan1}
\end{subfigure}%
\begin{subfigure}{.33\textwidth}
  \centering
  \includegraphics[width=5.5cm, height=5cm]{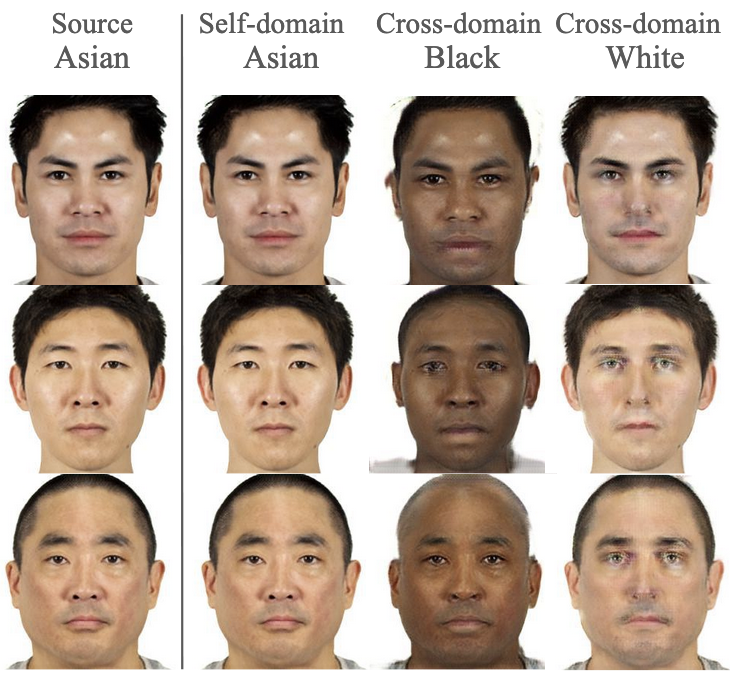}
  \caption{Asian to target domains}
  \label{racegan2}
\end{subfigure}%
\begin{subfigure}{.33\textwidth}
  \centering
  \includegraphics[width=5.5cm, height=5cm]{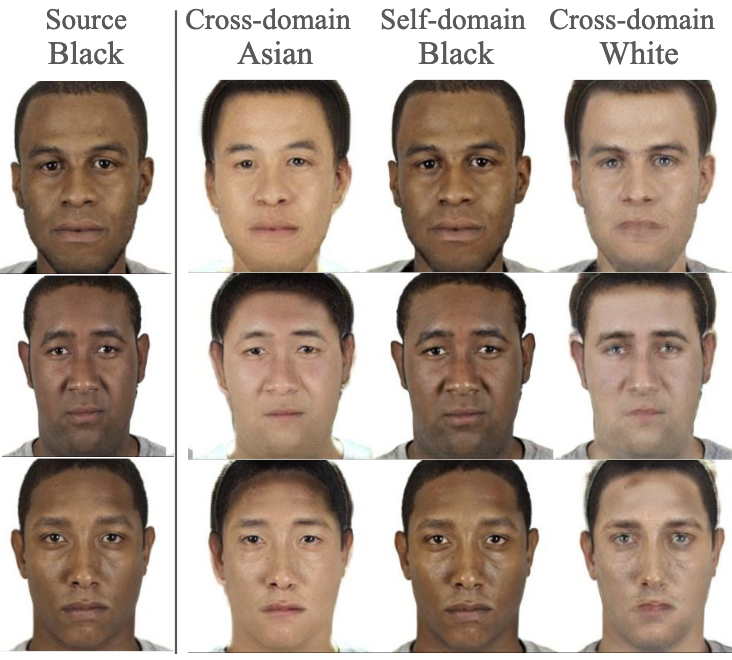}
  \caption{Black to target domains}
  \label{racegan3}
\end{subfigure}
\caption{Domain-style guided image synthesis with RaceGAN using the CFD dataset: (a) White to Asian, Black, and White, (b) Asian to Asian, Black, and White, and (c) Black to Asian, Black, and White.}
\label{fig:test}
\end{figure*}

\textbf{Style Extractor:} 
The input for the style extractor network is the target label $c$ and the latent code $z$. The latent code represents a Gaussian distribution sample. The extracted style code, $s$, is intended to be 256 (depending on the real image size), whereas the latent code's dimension is set to 16. With specific branches for every domain, the network primarily consists of a multilayer perceptron \cite{taud2018multilayer}. Both shared and unshared layers are part of the network's architecture. All domains have four completely linked shared layers, and each domain also has three fully connected unshared layers of its own. These layers have a set hidden dimension of 512. For each layer, the rectified linear unit (ReLU) is utilized as the activation function. An intermediate representation, $h$, is provided by the shared layers and is used to build domain-specific style code, $s$, by passing it via domain-specific unshared layers based on $c$. Custom mappings of unshared layers enable the style code to be more domain-specific. A thorough overview of this network is shown in Figure \ref{GMD}.

\textbf{Discriminator:} 
The discriminator network is composed of three layers, as shown in Figure \ref{GMD}: input, hidden, and output. For discriminator, no normalization is used. Before entering the hidden layer, the input images are convoluted by a $4\times4$ kernel of stride $2$. Leaky ReLU \cite{ramachandran2017searching} activation and five more layers of comparable utilities from the input layer make up the Hidden layer. The two layers that define the subsequent convolutional output layers are the classification layer, $D_{cls}$, which predicts the input domain, and the discriminating layer, $D_{src}$, which generates the probability distribution for the real or fake decision-making.

\begin{figure*}[!ht]
\centering
\begin{subfigure}{.5\textwidth}
  \centering
  \includegraphics[width=9cm,height=5cm]{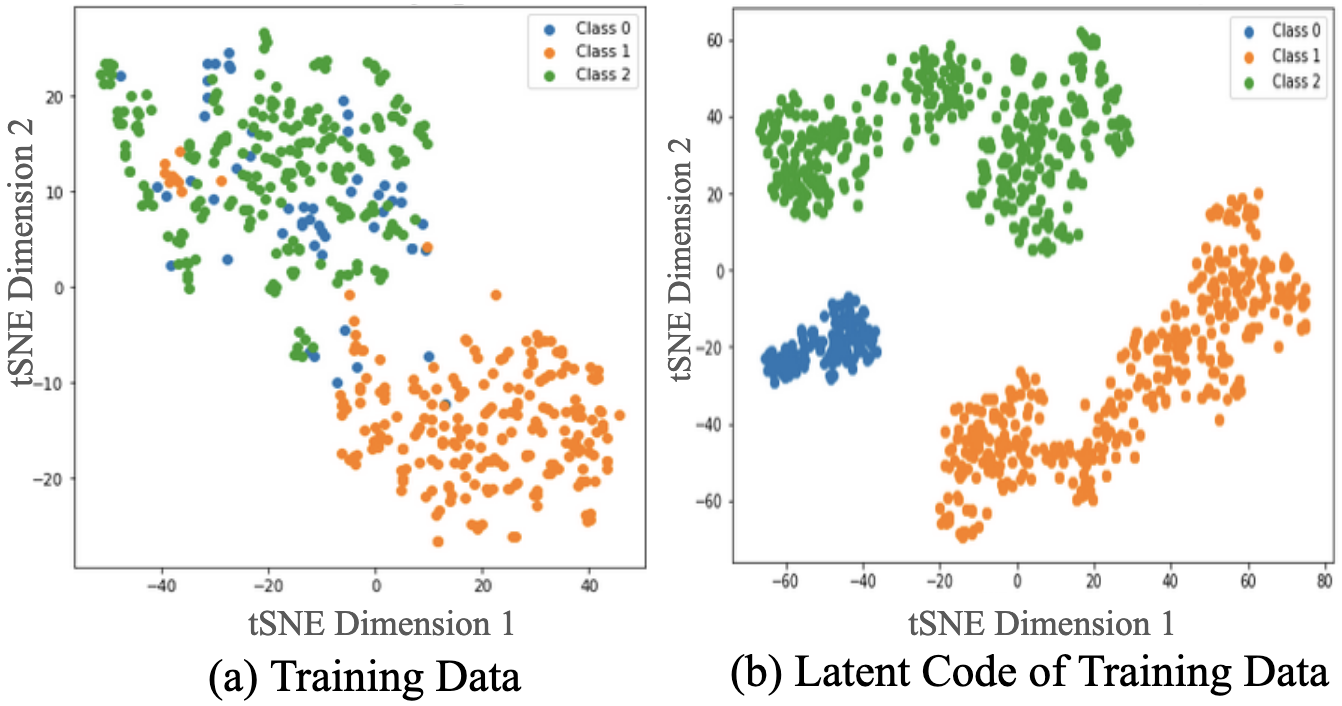}
\end{subfigure}%
\begin{subfigure}{.5\textwidth}
  \centering
  \includegraphics[width=9cm,height=5cm]{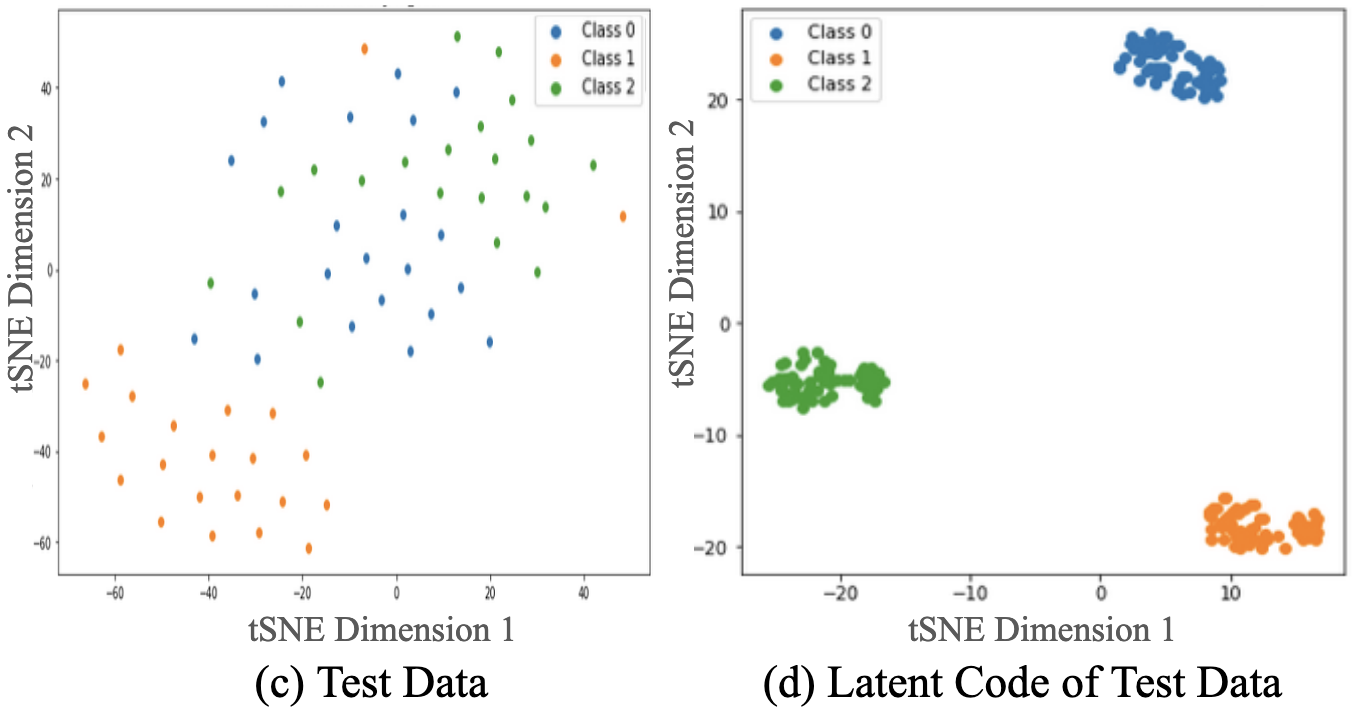}
\end{subfigure}%
\caption{t-SNE visualization of: (a) original distribution of training data; (b) latent code of training data generated from the encoder module of the generator; (c) original distribution of test data; (d) latent code of test data generated from the encoder module of the generator, where Class 0, Class 1, and Class 2 represent Asian, Black, and White classes, respectively.}
\label{tsne_tr}
\end{figure*}

\subsection{Training:}
We use data augmentation by employing random probability horizontal flipping, $1^{\circ}$ and $2^{\circ}$ random rotation. Due to the high resolution of the images, we isolate the most significant region center cropping at $1200\times1200 $limiting to the face. The images are then resized to $256\times256$. Because there aren't enough images in the dataset, we utilize 95\% of the data for training and randomly choose 5\% for testing. The number of batches is sixteen. Both the generator and the discriminator use an adaptable learning rate. For both, the starting learning rate is set at 0.0001, and it progressively decreases with each ${10}^{th}$ iteration. ADAM is used as an optimizer with $\beta_{1}=0.5$ and $\beta_{2}=0.9$. For improved performance, the model must train through at least 400,000 iterations. Model checkpoints and sample visuals are stored after every 1000 iterations.

\section{Experiments}
\subsection{Qualitative Evaluation:}
Considering all the specifications, we analyze other image-to-image translation models and compare the visual results with our model for qualitative assessment. Below is a list of the primary observations that are taken into account during the racial feature translation experiment:
\begin{itemize}
    \item Racial features (e.g., demographic skin color, eye size, ratio of upper and lower lips) should be translated.
    \item Pose and expression should be preserved.
    \item High-level semantics (e.g., makeup) should be preserved.
    \item Individual identities should be preserved. 
\end{itemize}

Some of the StarGANv2\cite{choi2020stargan} and StyleGAN2\cite{karras2019style} implementation's findings are shown in Figure \ref{img_compare}. StarGANv2\cite{choi2020stargan} is incredibly efficient at translating images with high quality. The generator, discriminator, mapping network, and style encoder are the four components that make up the model. Domain-specific style codes are retrieved from reference images using the style encoder. The generator uses these style codes in conjunction with the original image to produce domain-specific images that follow reference styles. Pairs of Asian and Black faces are utilized as reference images, while some multiracial faces are used as source, as Figure \ref{img_compare}\textcolor{blue}{a} illustrates. It is evident from analyzing the findings that some of the low-level semantics (like skin tone) and high-level semantics (like hairstyle) of the reference images were modified. Reversing the source and reference inputs in Figure \ref{img_compare}\textcolor{blue}{b} reveals a similar behavior. However, the synthesized images cannot be regarded as indicative of the source or the reference for any of the test instances. Later, we also explored StyleGAN2\cite{karras2019style} which is widely known for adopting progressive growth architecture to produce the most realistic, high-resolution images. This model does image synthesis guided by references. Figure \ref{img_compare}\textcolor{blue}{c} illustrates the experiment's visualization. It is clear from the illustration that the reference style has a significantly unpredictable influence on the synthesized images. The produced images' first column (highlighted in red) appears to have an incorrect style transfer. There has been no transfer of facial features or skin tone. The last synthesized image uses a Multiracial reference with a White source, but the resulting image depicts a Black person, which is inappropriate. 

Referring to Figures \ref{racegan1}, \ref{racegan2}, and \ref{racegan3}, we may assess how well our RaceGAN performed to meet the objectives. All these visuals, for comparison, comprise many source domain instances together with their corresponding cross-domain translated images and self-domain reconstruction images. It is evident from Figure \ref{racegan1} that White faces' racial traits are altered in order to translate them into the target cross-domains. For example, the skin tone changes to a yellowish tone when the target domain is Asian, making the eyelids less noticeable and puffier. All of the photos for the Black target domain feature a dark brown complexion, prominent jawbones, a prominent forehead, and fuller lips. Above all, each image retains its individuality and distinguishes itself from other images of the same race. However, only high-level characteristics and postures are preserved when manipulating racial features. Additionally, the White-to-White self-domain translation appears to be exact and suitable. Despite their lack of sharpness and resolution, these synthetic images accomplish the desired result. Figures \ref{racegan2} and \ref{racegan3} show similar phenomena. However, it should be noted that while the model does rather well when converting Black images to White, it has trouble producing flawless Asian faces. One possible explanation is that there is an imbalance in the dataset.

\subsection{Latent Space Exploration}
\begin{figure}[!t]
\vspace{-0.5cm}
    \centering
    \includegraphics[width=9cm,height=6.8cm]{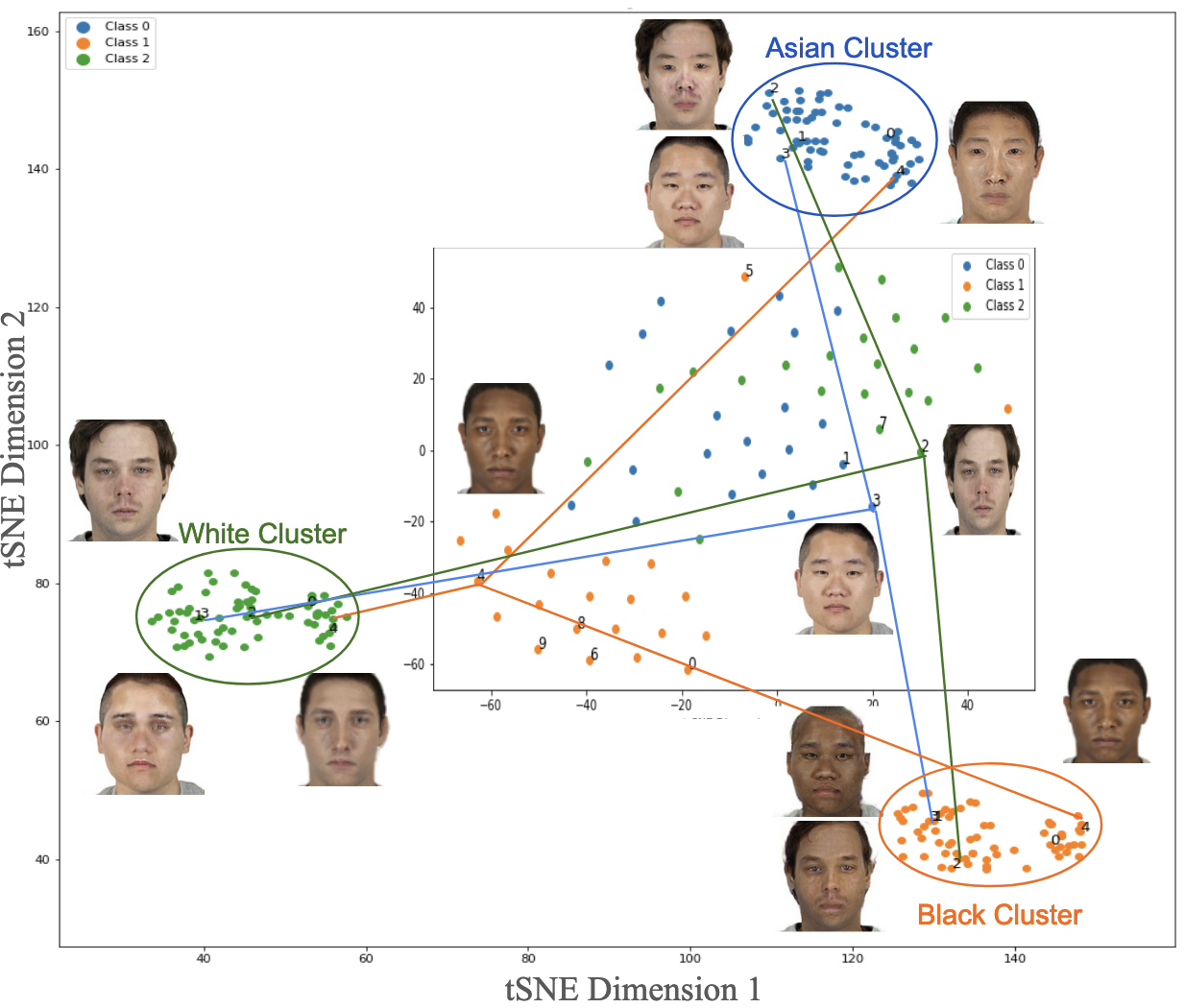}
    \caption{Visualization of original test data mapped in the trained latent space (center) and their respective translation for target domains. Note the clear clustering after translation.}
    \label{latent_mapping}
\end{figure}

For a better understanding of the model's performance, we tried to get insights from the latent space of the generator. We redesigned the model as an auto-encoder network and extracted latent code generated from the encoder. This latent code is used further by the decoder to reconstruct the converted images. We used a non-linear dimensionality reduction approach called t-SNE (t-distributed Stochastic Neighbor Embedding) \cite{hinton2002stochastic} for visualizing the high-dimensional data in 2D space. From Figure \ref{tsne_tr}, the original distribution and the latent distribution for both the training and testing sets can be visualized. Each point in the original distribution of Figures \ref{tsne_tr}\textcolor{blue}{a}, \ref{tsne_tr}\textcolor{blue}{c} represent training and testing images, whereas for the latent distribution in Figures \ref{tsne_tr}\textcolor{blue}{b}, \ref{tsne_tr}\textcolor{blue}{d}, the points indicate the original images and the other two race conversions representing one single point from the original distribution. The difference is easily noticeable in the latent distribution, where the model has perfectly clustered the latent codes into three clusters representing each class. Figure \ref{latent_mapping} shows the mapping of input images from the original data distribution to the latent distribution of target classes. As it can be clearly seen, the translated images are effectively clustered within their corresponding racial region in the latent space.

\subsection{Quantitative Evaluation:}
In order to provide a quantitative assessment, we employ a multi-class classifier that utilizes RaceGAN-generated  images and InceptionResNetv1 from facenet, which is specifically designed for face recognition \cite{schroff2015facenet}. Three experimental categories are used to complete the total evaluation: (a) training with self-domain translated images and testing with both self-domain and cross-domain translated images;  (b) training with self-domain translated images and testing with cross-domain translated images; (c) training and testing with cross-domain translated images. Cross-domain translation is the translation into other domains, such as Asian to White and Black, whereas self-domain translation is the translation into the input domain (i.e., Asian to Asian). We utilize self-domain translated images of size $256\times256$ instead of the high-resolution images from the dataset since it facilitates the model's comparison of features with test images of size $256\times256$. As we can see from Table \ref{classification}, all of the evaluation metrics are more than 92\% when we train the model using self-domain images and test it against images from both self-domain and cross-domain. Furthermore, even when testing against cross-domain translated images without the provision of self-domain translated images, this value retains up to 90\%. This demonstrates that the characteristics of our domain-translated images are similar to those of the actual images that correspond to certain domains. Additionally, when we train the model with only cross-domain translated images and test it with cross-domain images, the accuracy and other metrics climb to 98\%, indicating that the model is able to identify important features within cross-domain translated images.

\section{Conclusions}
In this paper, we propose RaceGAN, a multi-domain image-to-image translation model for racial attribute manipulation of human face images. The model is scalable for several target domains, requiring only a single generator and a single discriminator. It employs a style extractor module to extract domain-specific low-level style code from a target domain and fuse it with the input image of the source domain to generate racially translated images of the target domain, keeping high-level styles invariable. The model is also capable of separating racial domains in latent space, where each domain consists of the same set of images, either in original or converted formats. However during the experiment, male faces from just three prominent races have been investigated. We'll experiment with different racial groups including male and female faces in the future while trying to make the translation better.

\begin{table}[!t]

\caption{Comparison of the performance of race classification on translated images produced by RaceGAN. }
\label{classification}
\begin{tabular}{M{2.9cm}|M{0.9cm}M{1cm}M{0.7cm}M{1.2cm}}
\hline
\textbf{Experiment Setting} & \textbf{Accuracy} & \textbf{Precision} & \textbf{Recall} & \textbf{F1-score} \\ \hline
SSC & 92.4\% & 92.9\% & 92.7\% & 92.4\% \\\hline
SC & 88.9\% & 90.1\% & 88.9\% & 88.9\% \\\hline
CC & 98.3\% & 98.5\% & 98.3\% & 98.3\% \\ \hline
\end{tabular}
\footnotesize{SSC = Self-domain training - Self \& cross-domain testing, SC = Self-domain training - Cross-domain testing, CC = Cross-domain training \& testing}\vspace{-.5cm}
\end{table}
\section*{Acknowledgment}

This work was partially supported by the Department of Defense under grant number FA9550-21-1-0207, the National Institute of General Medical Sciences of the National Institutes of Health under grant number P30 GM145646 and by the National Science Foundation under grant number OAC 2201599.

\bibliographystyle{ieeetr}
\bibliography{refs}

\end{document}